\documentclass{article}

\usepackage{arxiv}
\usepackage[utf8]{inputenc}
\usepackage[T1]{fontenc}
\usepackage{hyperref}
\usepackage{url}
\usepackage{enumerate}
\usepackage{enumitem}
\usepackage{booktabs}
\usepackage{amsmath}
\usepackage{amsfonts}
\usepackage{nicefrac}
\usepackage{microtype}
\usepackage{graphicx}
\usepackage{natbib}
\usepackage{doi}
\usepackage{booktabs}

\title{Towards Automatic Continual Learning: A Self-Adaptive Framework for Continual Instruction Tuning}

\date{}

\author{
  Peiyi Lin\textsuperscript{1,2}, 
  Fukai Zhang\textsuperscript{1},
  Kai Niu\textsuperscript{2},
  Hao Fu\textsuperscript{1}\\[6pt]
  \textsuperscript{1}National Supercomputer Center in Tianjin, Tianjin, China \\
  \textsuperscript{2}School of Science, Tianjin University of Technology and Education, Tianjin, China
}

\renewcommand{\shorttitle}{}
\renewcommand{\undertitle}{}
\renewcommand{\headeright}{A preprint}

\begin{document}
\maketitle

\begin{abstract}
Continual instruction tuning enables large language models (LLMs) to learn incrementally while retaining past knowledge, whereas existing methods primarily focus on how to retain old knowledge rather than on selecting which new knowledge to learn. In domain-specific contexts, maintaining data quality and managing system constraints remain key challenges. To address these issues, we propose an automated continual instruction tuning framework that dynamically filters incoming data, which identify and reduce redundant data across successive updates. Our approach utilizes a small proxy model for efficient perplexity-based filtering, and updates the proxy to ensure that the filtering criteria remain aligned with the evolving state of the deployed model. Compared to existing static data selection methods, our framework can effectively handle incrementally acquired data and shifting distributions. Additionally, it addresses practical deployment challenges by enabling seamless model updates, supporting version rollback and incorporating automatic checkpoint evaluation. We evaluated the system in real-world medical scenarios. It reduced computational costs by 66.7\% and improved model performance, and achieved autonomous updates, thus demonstrating its effectiveness for automatic continual instruction tuning.
\end{abstract}

\section{Introduction}

Continual learning \citep{li2019learn} of large language models (LLMs) is a widely used technique, enabling model to adapt incrementally while minimize degradation on previously acquired knowledge. Nevertheless, applying continual instruction tuning to domain or task specific tasks introduces unique challenges that are not adequately addressed by general methods. In this paper, we focus on two key challenges: (1) ensuring data quality in a dynamic, real-world setting and (2) addressing system-level constraints in practical deployment.

One of the primary challenges in continual instruction tuning is ensuring data quality. The efficiency of instruction tuning heavily depends on the quality of the instruction data \citep{ren2024rethinking, zhou2023LIMA}, thereby low-quality data can result in issues such as catastrophic forgetting and overfitting. In the context of continual learning, domain-specific data often involves sensitive, undisclosed information, thereby acquiring high-quality data is both costly and labor-intensive. Although synthetic data helps overcome the scarcity problem, it doesn't always guarantee the quality of the data, as limited scenarios and tasks often result in redundant or duplicated information in the generated data.

To address these challenges, several frameworks have been proposed to select high-quality instruction data \citep{liu2024what, albalak2024survey}, mainly based on criteria like quality, diversity and necessity \citep{qin2025unleashing}. For instance, \citet{ge2024clustering} employs a judge model to rank data by quality and applies clustering algorithms to ensure diversity. \citet{du2023mods} selects data by evaluating model performance for quality and identifying a seed subset to enhance diversity. \citet{xia2024LESS} constructs a low-dimensional gradient datastore and selects data based on their similarity. Furthermore, other methods leverage model-derived metrics, such as perplexity, which measures a model's generative capabilities through probability. \citet{li2024quantity} propose a filter criteria called instruction-following difficulty (IFD) score calculated through perplexity, and perplexity-based criteria have also been widely applied in pre-training data selection \citep{thrush2024improving}. Despite their effectiveness, these methods typically depend heavily on model evaluation, thus making the selection procedure computationally expensive, especially when using large and capable models. To mitigate this limitation, \citet{li2024superfiltering} improves efficiency by using a smaller proxy model for calculating perplexity while preserving efficacy.

However, all the existing techniques primarily focus on static datasets and are not designed for continual instruction tuning. This setting requires handling incrementally acquired data, which means the new data is sequentially collected and integrated into the model over time. Current methods apply filtering criteria on data in isolation, considering only individual batches of data rather than the entire process. Additionally, these approaches remain static for each batch, failing to account for the dynamic nature of data across different stages of fine-tuning. Data characteristics such as quality and diversity can vary significantly before and after fine-tuning, therefore a fixed filtering strategy may not be effective throughout the process. As a result, existing techniques struggle to adapt to scenarios where data is updated over time or experiences a distribution shift. To overcome these limitations, it is essential to use a dynamic data filtering approach that continuously evaluates newly acquired data in relation to previously processed data.

In parallel, there are also several system-level challenges in the practical deployment of continual instruction tuning. In real-world scenarios, data privacy consideration is essential for sensitive domains or environments, which usually prefer local model deployment and updating incrementally acquired data in intranet. Moreover, updating model in these scenarios could meets additional constraints. For example, updating deployed LLMs without disturbing the existing inference service, and being able to promptly roll back to a previous version if the update does not perform well. While these factors could be achieved separately through existed framework, by manual operation or interrupted the service, it could be really cumbersome and not realistic to continually update it over time. Together, these conflicts call us to investigate this automatic framework for continually updating the data that been collected in cycles, which is suitable for the situation for practical deployment.

In this work, we design a iterative fine-tuning system to which automate the whole update process, including data filtering, continual instruction tuning, checkpoint evaluation and switching the deployed model. In particular, the automatic checkpoint evaluation provides a reliable quality of the tuning result and enables version control. In addition, this system also enables seamlessly update without stopping or restarting the deployment, which provide a more stable and reliable implement.

The proposed data filtering effectively manages incrementally acquired data using dynamic criteria. It employs a small proxy model for perplexity-based filtering, drawing inspiration from \citet{li2024superfiltering}. Moreover, we iteratively update this proxy model alongside updates to the deployed model. This approach ensures that the selection criteria remain aligned with the most recent updates, capturing dynamic changes in data distribution across iterations. This not only leads to a more accurate selection for the current model but also identifies redundant information learned from previous tuning. Compared to existing methods, the novelty of our framework lies in its automated and efficient management of incrementally acquired data within domain-specific continual instruction tuning. We validate our system in a real-world medical scenario, though the method is general enough to be applied to other domains as well. Experimental results demonstrate the system decrease 66.7\% of the amount of the training data and computational cost compares to using full dataset, and get a better result.

To summaries, our work makes the following key contributions:
\vspace{-5pt}
\begin{itemize}[leftmargin=15pt]
    \item We develop a complete automatic system that supports domain continual learning, ensuring seamless model updates without service interruptions.
    \item We introduce a self-adaptive data filtering framework that continuously evaluates newly acquired data based on its relevance to previous updates, which remains effective even as data distribution evolves over time.
    \item We validate our framework in a real-world medical domain, demonstrating the framework reduces computational costs by 66.7\% and improved model performance compared to using the full dataset without manual operation.
\end{itemize}

\section{Related Work}
\textbf{Continual learning.} Continual learning \citep{li2019learn} focuses on developing learning algorithms to accumulate knowledge on a continual data stream, while tackling the catastrophic forgetting of old knowledge. The traditional methods can be roughly classified into three groups: rehearsal, regularization, and architectural, along with some hybrid techniques \citep{biesialska2020continual}. Regularization-based methods, such as elastic weight consolidation (EWC) \citep{kirkpatrick2017ewc} and synaptic intelligence (SI) \citep{zenke2017synaptic}, modify the ways in how parameters are updated. Rehearsal-based methods, such as experience replay \citep{chaudhry2019tiny} and generative replay \citep{shin2017continual}, retrain the model with previously seen data. Additionally, dynamic architecture methods like progressive networks (PNN) \citep{rusu2016progressive}, modify the model’s structure to prevent forgetting. While some studies explore data selection, they primarily focus on optimizing memory buffers rather than leveraging the information within the data itself. Different from traditional continual learning methods, which prioritize mechanisms to preserve past knowledge, our approach takes a different perspective: instead of focusing solely on retention, we improve data quality for new knowledge to directly reduce the likelihood of forgetting at its source. 

\textbf{Instruction Tuning Data Selection.} Studies have shown that data selection is both necessary and effective for instruction tuning \citep{liu2024what, albalak2024survey}. Research by \citet{ren2024rethinking, zhou2023LIMA} suggests that most of an LLM's knowledge is derived from pretraining, and a limited amount of high-quality data is sufficient for achieving strong performance. Existing frameworks mainly focus on three key aspects: quality, diversity, and necessity, along with some hybrid methods \citep{qin2025unleashing}. Diversity selection is often performed by unsupervised clustering algorithms \citep{ge2024clustering} or the initial selection of a subset of seed data \citep{du2023mods}. Necessity selection estimates the impact of specific data samples using gradient-based methods \citep{xie2023resampling} or influence functions\citep{xia2024LESS}. For quality selection, model-based evaluators are commonly used \citep{zhou2023LIMA}, while perplexity has also been explored as a model-based criterion \citep{li2024quantity, li2024superfiltering}. One major advantage of using perplexity-based criteria is that it enables importance and quality measurement at the same time, since it can filter out some irrelevant data based on the absolute perplexity values and assess importance through model gradient analysis. In this paper, we propose an integrated approach that aims to achieve all three aspects by relying solely on perplexity calculations. Specifically, diversity is ensured by identifying redundant information within the dataset across different updates through the dynamic updating of the proxy model.

\textbf{Medical Instruction Tuning Dataset.} We utilize a medical consultation task with a generated synthetic dataset to evaluate our system, which is an extensively explored scenario in both research and practical applications. Existing studies have proposed several medical-specific natural language processing (NLP) datasets and approaches for fine-tuning medical LLMs. CBLUE \citep{zhang2022cblue} contains eight annotated biomedical language understanding tasks datasets annotated by doctors, while BigBio \citep{fries2022bigbio} provides a framework for the standardized biomedical NLP meta-dataset generation. For multi-turn medical dialogues, HealthCareMagic-100k \citep{li2023chatdoctor} and Zhongjing \citep{yang2024zhongjing} have both constructed datasets derived from real doctor-patient consultations. In addition to human experts, several studies have leveraged automatic methods like medical knowledge graphs to ensure high-quality generated instruction data. HuatuoGPT \citep{zhang2023huatuogpt} and MedChatZH \citep{tan2024medchatzh} generate dataset based on domain-specific knowledge while filtering out irrelevant or sensitive content. DiscMedLLM \citep{bao2023discmedllm} combines different types of data with knowledge graph-driven sample construction, reconstructed real-life dialogue, and modeling of human preferences. Moreover, BianQue \citep{chen2023bianque} focuses on refining medical dialogue interactions through a structured chain of questioning (CoQ) framework.

\section{System Design}
\begin{figure}[htbp]
    \centering
    \includegraphics[width=\textwidth]{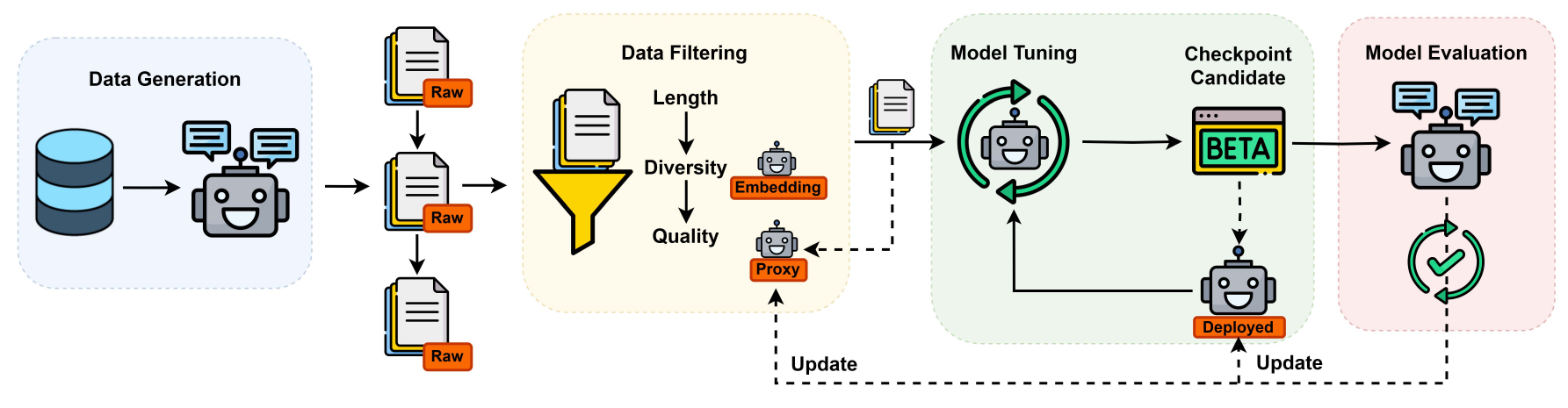}
    \caption{An illustration of the full framework. The data generation module utilizes the Chain-of-Thought (CoT) approach to generate synthetic data when the dataset contains only instructions. The data filtering module processes data based on length, diversity, and quality as a criterion. The model tuning module updates the currently deployed model, producing a checkpoint candidate, which is then assessed by the model evaluation module. If the candidate outperforms the current model, both the deployed model and the proxy model used for quality measurement are iteratively updated.}
    \label{workflow}
\end{figure}

Our proposed framework consists of four modules: data generation, data filtering, model tuning and model evaluation. The overall structure is illustrated in Figure~\ref{workflow}.

\subsection{Problem Setup}
In a continual instruction tuning setting \citep{biesialska2020continual}, the model learns from a sequence of tasks \(  T= \{T_1, T_2, ..., T_n\} \), where each task \( T_i \) is associated with a dataset \( \mathcal{D}_i \) consisting of instruction-response pairs: 

\begin{equation}
    \mathcal{D}_i = \{(x_{ij}, y_{ij}) \mid j = 1, ..., N_i \}
\end{equation}

where \( x_{ij} \) is the \( j \)-th instruction in task \( T_i \), and \( y_{ij} \) is its corresponding response. The total number of instruction-response pairs in task \( T_i \) is denoted by \( N_i \). At each step \( i \), the model is trained on \( T_i \) without access to previous datasets \( \mathcal{D}_1, ..., \mathcal{D}_{i-1} \). The goal is to optimize model performance on \( T_i \) while preserving knowledge from prior tasks and efficiently adapting to new ones.

\subsection{Data Generation}
\label{data}
We constructed a high-quality medical instruction tuning dataset from real-world medical record, which contains patient symptom, past medical history, and auxiliary examination results. We guided the model to generate question-answer pairs for various medical consultation scenarios, including symptom identification, prescription recommendations, and medical examinations. To ensure alignment with professional medical standards, we explicitly implemented several constraints several constraints: the model was required to adopt the role of a medical expert; detailed explanations were provided for constructing responses; the inclusion of false information was strictly prohibited; answers had to exceed a certain length to provide sufficient detail; and all outputs needed to conform to a structured JSON format.

A key aspect of our approach was integrating Chain-of-Thought (CoT) \citep{WeiS2022cot} prompting into the dataset to enhance the interpretability and reliability of the model’s diagnostic process. We designed prompts that required the model to simulate a doctor’s reasoning steps when diagnosing, in order to leveraging the language understanding and generation capabilities of LLMs. Specifically, the model was instructed to generate not only diagnostic results but also supporting diagnostic evidence, to ensure that each response was grounded in a reasoning process. This allowed us to systematically extract diagnostic evidence from raw medical records, thereby constructing a dataset that incorporates explicit reasoning chains for diagnostics.

Using this approach, we generated a Chinese medical fine-instruction tuning dataset, consisting of structured question-answer pairs along with step-by-step diagnostic reasoning. This dataset improves the model’s ability to perform medical consultation tasks by improving interpretability, and addressing hallucination issues by eliminating examples that contain incorrect information from the original records. By integrating real medical records, enforcing strict instruction constraints, and CoT prompting in the dataset, we successfully conducted high-quality medical responses that are evidence-based and include clear diagnostic reasoning chains.

\subsection{Data Filtering}
We applied a data filtering strategy to improve the quality of training data based on the properties of the response \( y_{ij} \). Specifically, the initial preprocessing would focus on two criteria: response length and semantic diversity, which is calculation-effective and eliminate low-quality samples. In addition, a small proxy model with fewer parameters is used for calculating perplexity, which measures how necessary the instruction would be to the model, helping reduce unnecessary samples for the instruction tuning process. Finally, the proxy model would be iteratively updated when the deployed model is updated to dynamically align the criteria over time, and enable the identification of redundant data samples that the model has already learnt the knowledge in previous terms. This data filtering module ensures that only coherent and high-quality data are considered for further training.

\subsubsection{Length-Based Preprocessing} 
Longer responses often provide more contextual information, while very short responses may lack sufficient content. For each response \( y_{ij} \) in the dataset \( \mathcal{D}_i \), the length \( l_{ij} \) is measured by the number of sentences. In most cases, we prefer to retain longer responses, so responses shorter than a predefined threshold \( L_{\text{min}} \) are removed: 

\begin{equation}
\mathcal{D}_{L_i} = \{ (x_{ij}, y_{ij}) \in \mathcal{D}_i \mid l_{ij} \geq L_{\text{min}} \}
\end{equation}

where \( \mathcal{D}_{L_i} \) denotes the subset of data from \( T_i \) that satisfies the length criterion.  

\subsubsection{Semantic Diversity-Based Preprocessing}  
To ensure that responses are semantically diverse and sufficiently informative, we compute a semantic diversity score for each response \( y_{ij} \) using pairwise cosine similarity between its sentences. For a given response \( y_{ij} \) consisting of \( m \) sentences \( s_1, s_2, ..., s_m \), the semantic diversity score \( s_{ij} \) is computed as follows:\par
\hspace{15pt}\textbf{Step 1: }Each sentence \( s_k \) is embedded using a pre-trained language model. Experimentally, we found sentence-level embeddings to be more effective than token-level embeddings.\par
\hspace{15pt}\textbf{Step 2: }Pairwise cosine similarity is computed between all sentence embeddings within \( y_{ij} \).\par
\hspace{15pt}\textbf{Step 3: }The diversity score is calculated as:
    
\begin{equation}
s_{ij} = 1 - \frac{1}{m(m-1)} \sum_{k=1}^{m} \sum_{l=k+1}^{m} \cos(\mathbf{h}_{s_k}, \mathbf{h}_{s_l})
\end{equation}

where \( \mathbf{h}_{s_k} \) and \( \mathbf{h}_{s_l} \) represent the embeddings of sentences \( s_k \) and \( s_l \), respectively. A higher \( s_{ij} \) indicates greater semantic diversity, while a lower score suggests uninformative.

After calculating the semantic diversity scores, we filter the data accordingly. Responses with a semantic diversity score below a predefined threshold \( S_{\text{min}} \) are filtered, ensuring that only qualified responses are retained:  

\begin{equation}
    \mathcal{D}_{S_i} = \{ (x_{ij}, y_{ij}) \in \mathcal{D}_{L_i} \mid s_{ij} \geq S_{\text{min}} \}
\end{equation}

where \( \mathcal{D}_{S_i} \) represents the set of instruction-response pairs that satisfy both length and semantic diversity criteria for task \( T_i \).

\subsubsection{Perplexity-Based Filtering}
After deal with the noisy data through the previous preprocessing, we then employ a filtering process based on Instruction-Following Difficulty (IFD) proposed by \citet{li2024quantity} as our final process, which quantifies the necessity of the instruction for generating the response through perplexity (PPL). These metrics allow us to filter out noisy, low-quality, or instruction-irrelevant samples.

Perplexity is a widely used metric for assessing the quality of a generated response by measuring how well a language model predicts the next token given its preceding context. A lower perplexity value indicates that the response is more fluent and predictable, while a higher value suggests incoherent or low-quality text. For a given response \( y_{ij} \) consisting of \( n_{ij} \) tokens, its perplexity is computed as:

\begin{equation}
PPL(y_{ij}) = \exp \left( -\frac{1}{n_{ij}} \sum_{k=1}^{n_{ij}} \log P(y_{ij,k} \mid y_{ij,<k}) \right),
\end{equation}

where \( P(y_{ij,k} \mid y_{ij,<k}) \) denotes the conditional probability of the \( k \)-th token given its preceding tokens.

The IFD score is calculated through the ratio of response perplexities in different conditions, quantify the importance of the instruction \( x_{ij} \) in generating the response \( y_{ij} \). To be specific, \( PPL(y_{ij} \mid x_{ij}) \) represent the perplexity when the model generates \( y_{ij} \) conditioned on the instruction \( x_{ij} \), while \( PPL(y_{ij})\) is the perplexity when the model generates \( y_{ij} \) without the instruction, indicate the model's original capability without instruction.

\begin{equation}
IFD_{ij} = \frac{PPL(y_{ij} \mid x_{ij})}{PPL(y_{ij})}
\end{equation}

Generally, a higher \( IFD_{ij} \) score indicates that the response is significantly influenced by the instruction, suggesting that the instruction provides necessary guidance for response generation. In contrast, a lower \( IFD_{ij} \) score implies that the response can be generated with little reliance on the instruction, which may indicate generic, redundant, or instruction-irrelevant data. It is noticeable that a IFD score above 1 or below 0 would be view as anomaly, imply that instructions provide minimal or no benefit, represent to instruction-irrelevant, or lower-quality data samples.

To retain only meaningful instruction-following examples, we define a minimum IFD threshold \( IFD_{\text{min}} \) and filter out samples. Thus, the final dataset after applying both perplexity and IFD filtering is:

\begin{equation}
\mathcal{D}_{IFD_i} = \{ (x_{ij}, y_{ij}) \in \mathcal{D}_{S_i} \mid 1 > IFD_{ij} \geq IFD_{\text{min}} > 0\}.
\end{equation}

By filtering based on IFD, this approach ensures that the dataset consists of high-quality, instruction-relevant examples, thereby improving the model’s ability to generalize across new tasks in a continual learning setting. improving its generalization ability across tasks in the continual instruction tuning process.  

\subsubsection{Proxy Model Update}
Previous work by \citet{li2024superfiltering} improved the calculation of IFD scores by utilizing a small proxy model, demonstrating its effectiveness in comparison to directly employing larger LLMs. However, this approach still does not fully address the challenges inherent in continual learning scenarios, which require adaptive criteria to manage shifting data distributions over time.

\begin{figure}[htbp]
    \centering
    \includegraphics[width=\textwidth]{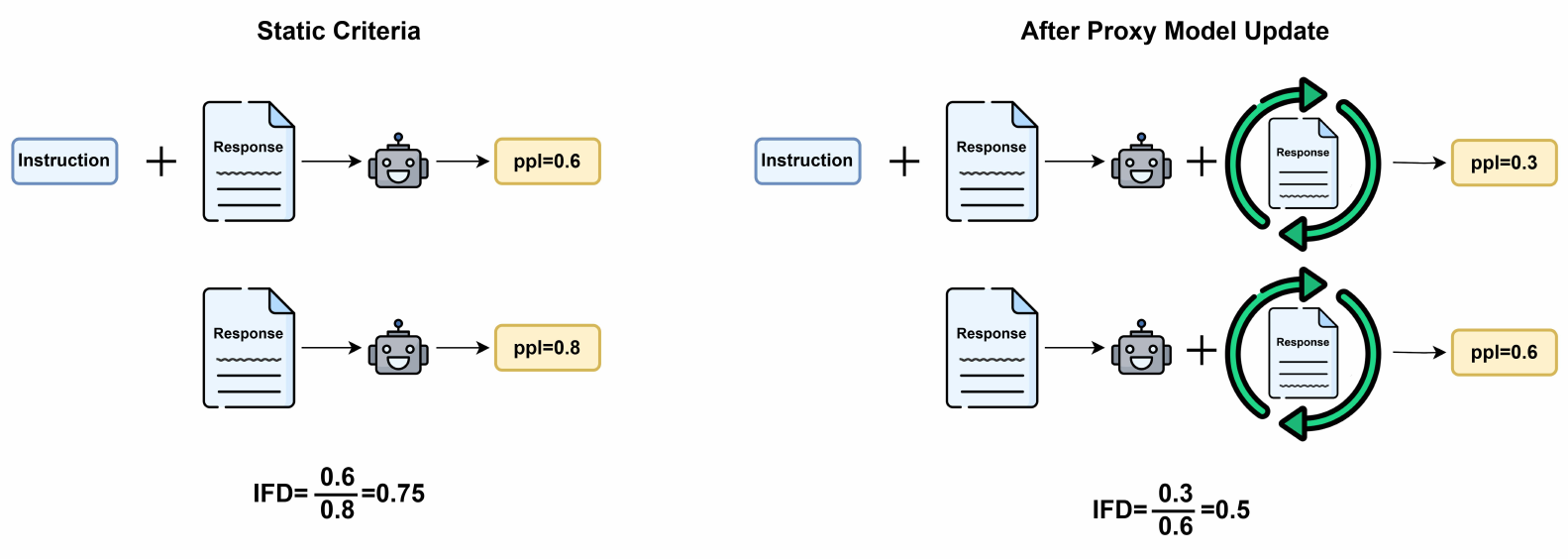}
    \caption{An illustration of the dynamic criteria. The proxy model used for calculating perplexity would yield a different output after tuning, resulting in the identification of redundant information.}
    \label{update}
\end{figure}

To address this limitation, we propose iteratively updating the small proxy model alongside the deployed large model, continual instruction tuning both simultaneously on the same data. As the updates progress, the proxy model evolves continuously, enabling it to accurately reflect the current state and challenges faced by the deployed model. This iterative process work in tandem with the update and model evaluations, creating a feedback loop within the whole system, allowing IFD scores computed by the proxy model remain relevant and accurate despite shifts in data distribution or changes in task requirements. The calculation and update process is illustrated in Figure~\ref{update}.

Synchronized update to the proxy model enhance its sensitivity and relevance, enabling it to estimate IFD scores more accurately, which offers several benefits:

Firstly, as the proxy model learns from the data, it dynamically adjusts its internal criteria for identifying challenging examples. This adaptive filtering method helps the deployed model maintain focus on incrementally acquired data across various update cycles, promoting sustained improvement, which is especially beneficial in domain-specific continual learning.

Secondly, continuously updating the proxy model aids in identifying redundant information that the deployed model has already learned from previous updates. Filtering out these redundant samples reduces unnecessary computational overhead and mitigates the risk of overfitting.

Finally, an iteratively updated proxy model can naturally adapt to new data distributions, regardless of whether they differ slightly or significantly from previously encountered data. This adaptability ensures robust performance and enhances the overall effectiveness of the deployed large model.

By implementing this iterative co-update approach, we ensure that the data selection process remains high-quality and closely aligned with the evolving needs of both the deployed model and the continuously shifting data distribution, ultimately leading to improved continual learning performance.

\subsection{Model Tuning}
The tuning module utilizes continuous instruction tuning of the deployed LLM. To be specific, it fine-tunes the most recently deployed model checkpoint with the filtered instruction data from the data filtering, and generates a candidate checkpoint. This candidate checkpoint is subsequently evaluated by the model evaluation to determine its improvement over the currently deployed model. If the evaluation indicates superior performance compared to the currently deployed checkpoint, the tuning module will load the existing checkpoint instead of the improved candidate checkpoint for the upcoming inference service and refresh the initial update state for the next iteration. 

We employ Parameter-Efficient Fine-Tuning (PEFT) methods, specifically Low-Rank Adaptation (LoRA) \citep{hu2022lora}. This approach significantly reduces computational costs by updating only a subset of parameters rather than the entire model. While preserving most of the knowledge from the original model, the risk of overwriting historical knowledge has also been minimized, making it suitable for continual learning \citep{biderman2024lora}. Furthermore, the lightweight characteristics of LoRA weights decrease the size of the tuned checkpoint, enabling fast and seamless update without the need to reload the entire model.

\subsection{Model Evaluation}
The model evaluation module systematically validate the performance of each newly tuned candidate checkpoint against the currently deployed checkpoint. Evaluation can be conducted in two ways: direct accuracy evaluation and LLM-based evaluation, depending on the availability of a validation dataset.

On one hand, when a validation dataset with explicit answers is available, direct quantitative evaluation using accuracy metrics is performed to objectively measure the candidate checkpoint's performance improvements.

On the other hand, in scenarios that only include validation instructions or lack explicit answers in response, we utilized LLM-as-a-judge method \citep{zheng2023judging}. Although it is more costly than other scenarios, it offers a qualitative and automatic method for unsupervised datasets. The judge LLM compares the responses generated by the candidate checkpoint with those from the currently deployed model, providing qualitative comparative assessments.

The model evaluation module offers feedback to the tuning module, indicating whether the candidate checkpoint represents an improvement over the currently deployed model. Positive evaluation results trigger updates in both the data filtering and tuning modules. In contrast, negative evaluation results lead directly to the next cycle without taking actions.

\subsection{Iterative Automatic Update Cycle}
The proposed four-module framework operates iteratively without manual intervention. Each iteration begins with data generation and filtering, continues with continual instruction tuning, and concludes with automatic evaluation. When evaluation results confirm positive performance gains, the deployed inference model, initial update state, and proxy model are all updated accordingly. This cyclical framework ensures the continuous improvement of the deployed LLM, maintains data efficiency, and adaptively responds to incoming data over time.

\section{Experiments}
\subsection{Experiment Setting}
\textbf{Datasets.} We use the generated dataset in Section~\ref{data} for the dataset of instruction tuning with the amount of 30000, where the training set and validation set are divided in a 9:1 ratio. The test set is written by human experts in the same format for checkpoint evaluation, which focuses on symptom identification and prescription recommendations. We also validate our system on a medical benchmark IMCS \citep{chen2022IMCS} from MedBench \citep{cai2024MedBench}, which is a rare and valuable open-source Chinese medical domain dataset. It is a dataset that contains medical dialogue from different tasks. We focus on judging diagnosis, which is clear to evaluate the result.

We utilized the general task dataset from \citet{ji2023exploring} to reduce the overfitting problem in domain-specific tuning. In particular, we applied data filtering to this dataset and randomly selected samples from it, mixing them with the professional dataset in a 1:1 ratio, which has been experimentally shown to be effective.

\textbf{Evaluation Metrics.} We use accuracy to evaluate the model’s performance in disease diagnosis of the generated dataset by exact disease name matching. A prediction is considered correct if the model provides a diagnosis that exactly matches the ground truth. Predictions that fail to match the ground truth are marked as wrong. Additionally, it would be viewed as a fault case if the model generates a prediction that isn't a diagnosis or not in the required format. Accuracy is computed as the ratio of correct cases to the total cases.

For the IMCS dataset, BLEU and ROUGE-L are used as evaluation metrics for the whole diagnosis. As this dataset does not adhere to a strict diagnostic format and requires a more flexible evaluation that considers both the diagnosis and the supporting evidence. BLEU measures the overlap of n-grams between the model’s output and reference texts, capturing lexical similarity, while ROUGE-L evaluates the longest common subsequence, reflecting the relevance and completeness of the diagnosis evidence.

\textbf{Model parameters.}
In our instruction tuning experiments, we fine-tuned the Qwen2.5-14B-Instruct model in BF16 precision. We used a training batch size of 4 per device, an evaluation batch size of 1 and accumulated gradients over every 8 steps. For optimization, we chose the AdamW optimizer, with a learning rate at 2e-5, weight decay at 0.03, and beta2 at 0.95. We also implemented a cosine learning rate scheduler with a warmup ratio of 0.1. Additionally, LoRA is applied with rank 16, alpha 32, and dropout rate of 0.05. To speed up training, we enabled DeepSpeed ZeRO Stage 3 and Flash Attention 2. Moreover, we set the maximum sequence length to 4096 and used lazy dataset loading.

In the evaluation settings, we used the same model for inference in BF16 precision. The maximum sequence length was set to 1024 tokens, with the number of newly generated tokens at 128. We enabled sampling-based generation without using beam search. For sampling parameters, we set the temperature to 0.7 and used a top-p value of 0.9. 

\textbf{Experiment Details.} 
We simulate continual data updates by dividing 3 million real-world generated data samples into 5 batches. The thresholds in the data filtering are set based on our generated data distribution as: minimum length \(L_{\text{min}}=800\), minimum diversity score \(S_{\text{min}}=0.5\), and minimum IFD score \(IFD_{\text{min}}=0.6\), which are select based on the data distribution. We utilized the all-MiniLM-L6-v2 as the embedding model for calculating the diversity score is all-MiniLM-L6-v2, while the Qwen2.5-0.5B-Instruct is used for calculating perplexity for IFD score. The experiments were conducted on a server with an Intel Xeon Silver 4314 CPU (2.40 GHz), two Nvidia Tesla A800 80GB GPUs, and 392GB of RAM.

\subsection{Main Results}
\textbf{Effectiveness of the Data Filtering.} To assess the performance of our data filtering, we conducted experiments using the generated medical training dataset with 30,000 samples and a test set of 429 cases. We applied the filtering method to select subsets of 10,000 samples in the order of higher IFD score, and compared their performance against the full dataset. 
\begin{table}[htbp]
    \centering
    \caption{Model performance after instruction tuning across different sample sizes.}
    \label{experiment 1}
    \begin{tabular}{@{}ccccc@{}}
    \toprule
    \textbf{Sample Size}   & \textbf{Accuracy} & \textbf{Correct Cases} & \textbf{Wrong Cases} & \textbf{Fault Cases} \\ \midrule
    0 (Before Tuning)      & 67.6\%             & 290                    & 104                  & 35                   \\
    10,000 (Filtered Data) & 70.4\%             & 302                    & 106                  & 21                   \\
    30,000 (Full Data)     & 70.2\%             & 301                    & 112                  & 16                   \\ \bottomrule
    \end{tabular}
\end{table}

The result of Table~\ref{experiment 1} shows both the filtered dataset and full dataset tuning result in a better performance, improving the accuracy for 2.8\% and 2.6\%, respectively. The filtered dataset of 10,000 samples dataset outperformed the full dataset, achieving a slightly higher accuracy despite using only one-third of the data and computational cost. Fault cases were reduced from 35 of baseline to 21, indicate the filtered data helped the model generalize better and avoid meaningless or incorrect outputs. This proved that through only letting model learnt the necessary data not only a effective way, but also reduce risk of overfitting through carefully selected.

While the full dataset marginally reduced fault cases compared to the filtered dataset, it also increased wrong cases. The slight decrease in accuracy suggests that training on excessive, unfiltered data introduces noise, which may lead to overfitting or less effective learning. Since the filtered dataset achieved better performance with significantly fewer samples, this confirms that not all additional data contributes positively.

In this case, the data filter significantly reduces the computational cost by 66.7\% while still achieving the best accuracy, demonstrating successful selection for high-quality samples and avoiding unnecessary updates. This is especially beneficial for practical continual learning scenarios where model updates need to be frequent but efficient.

\textbf{Dynamic Adaptation for Evolving Data.} Our experiments reveal noticeable differences in the IFD scores for a subset of samples when using the dynamically updated small LLM compared to the static version. Specifically, some of the samples that initially exhibited moderate difficulty under the static model showed decreased IFD scores after the proxy model was updated, suggesting that the updated model had already learnt this knowledge from the previous data before.

The refined IFD scores from the dynamic proxy model led to an identification of redundant information across samples from different updates. The finding suggests that some samples, previously overrepresented in training with a static proxy, are now being effectively leveraged. While we are still doing full quantitative analyses, early signs show that aligning the proxy model with the changing data distribution enhances the ongoing learning effectiveness.

\section{Conclusion}
In this work, we introduce an automated continual instruction tuning framework aim to address data quality problem and system constraints. Our proposed approach selects high-quality data before learning, thereby mitigating the risk of catastrophic forgetting and enhancing overall efficiency. By leveraging a dynamic data filter from a small proxy model, our method effectively reduces redundant data and avoids unnecessary updates. Furthermore, the system addresses practical deployment constraints by enabling automatic checkpoint evaluation and seamless updates without service disruptions. This provides a automatic and cost-effective solution for practical deployment in continual learning scenarios, particularly for domain-specific or task-specific applications.

\section{Future Work}
We plan to leverage some of the filtered-out data for further improvement. Although we initially filtered out these data samples due to their overlapping information with existing tuning results, they still maintain high quality individually. A practical approach might involve using them differently, such as for model alignment through Direct Preference Optimization (DPO) \citep{rafailov2023dpo}. Specifically, we propose employing a small-parameter LLM to generate synthetic responses that can serve as rejection responses for some of the discarded instruction data. However, this procedure must be carefully designed to avoid introducing biases or misleading patterns from the data. In future work, we will explore strategies to effectively balance alignment and instruction tuning in continual learning.

Future improvements may also explore how this framework can be effectively integrated with traditional rehearsal or regularization methods. Furthermore, it would be valuable to investigate a more elegant and adaptive filtering criterion to further optimize data selection over time.

\bibliographystyle{unsrtnat}

\end{document}